\documentclass{article}

 \usepackage[preprint]{neurips_2026}

\usepackage[utf8]{inputenc} 
\usepackage[T1]{fontenc}    
\usepackage{hyperref}       
\usepackage{url}            
\usepackage{booktabs}       
\usepackage{amsfonts}       
\usepackage{nicefrac}       
\usepackage{microtype}      
\usepackage{xcolor}         
\usepackage{caption}
\usepackage{float}
\usepackage{placeins}

\usepackage{booktabs}
\usepackage{multirow}
\usepackage{array}
\usepackage[table]{xcolor}
\usepackage{colortbl}
\usepackage{tabularx}
\usepackage{makecell}
\usepackage{graphicx} 
\usepackage{amsmath} 
\usepackage{threeparttable}
\usepackage{pifont}
\usepackage{fontawesome5}
\newcommand{\cmark}{\color{green}{\ding{51}}}%
\newcommand{\xmark}{\color{red}{\ding{55}}}%
\newcommand{\spectral}{\raisebox{-0.1em}{\includegraphics[width=3em,height=0.8em]{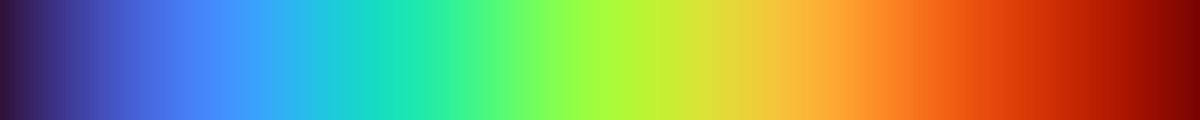}}}

\title{Unlocking Dense Metric Depth Estimation in VLMs}

\author{
    Hanxun Yu$^{1,2}$\footnotemark[1] \,\footnotemark[2]\quad
    Xuan Qu$^{1,2}$\footnotemark[2]\quad
    Yuxin Wang$^{2,3}$\quad
    Jianke Zhu$^{1,4}$\quad
    Lei Ke$^{2}$\quad 
    \\ \\
    $^{1}$Zhejiang University \quad $^{2}$Tencent Hunyuan LLM \quad $^{3}$HKUST \quad $^{4}$Shenzhen Loop Area Institute\\
    \\
    \faGithub \ \textbf{Project Page:} \href{https://depthvlm.github.io/}{https://depthvlm.github.io/}
}

\begin{document}


\maketitle

\renewcommand{\thefootnote}{\fnsymbol{footnote}}
\footnotetext[1]{Work done during an internship at Tencent Hunyuan LLM.}
\footnotetext[2]{Equal contribution.}

\begin{figure}[h]
\centering
\vspace{-5mm}
\includegraphics[width=1.0\linewidth]{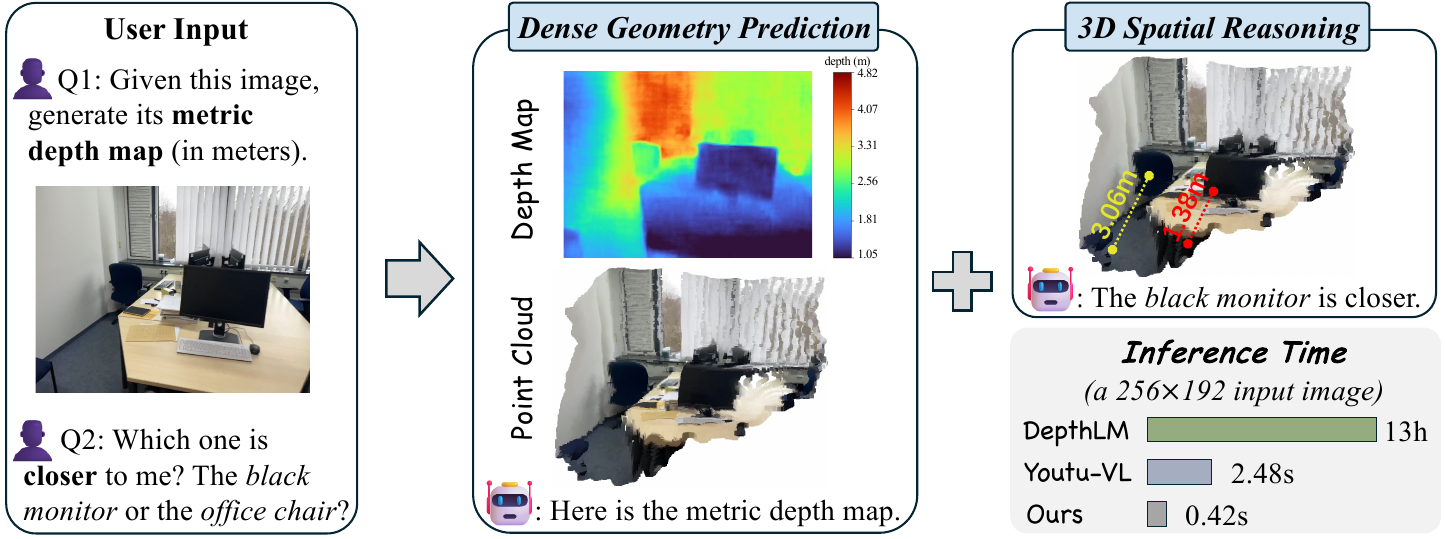}
\vspace{-4mm}
\caption{Our method serves as a unified foundation model for both low-level dense geometry prediction and high-level multimodal understanding, while achieving substantially faster inference compared with existing VLM-based approaches such as DepthLM~\cite{cai2025depthlm} and Youtu-VL~\cite{wei2026youtu-vl}.}
\label{teaser1}
\end{figure}

\begin{abstract}
Vision–Language Models (VLMs) excel at 2D tasks such as grounding and captioning, yet remain limited in 3D understanding. A key limitation is their text-only supervision paradigm, which under-constrains fine-grained visual perception and prevents the recovery of dense geometry. Prior methods either distill geometry from external vision models, introducing error accumulation, or enable direct prediction with inefficient per-pixel query or coarse token-level outputs. In this paper, we propose \textbf{DepthVLM}, a simple yet effective framework that transforms a single VLM into a \emph{native dense geometry predictor} while preserving its multimodal capability. By attaching a lightweight depth head to the LLM backbone and training under a unified vision–text supervision paradigm with a two-stage schedule, DepthVLM generates full-resolution depth maps alongside language outputs in a single forward pass. We further introduce a unified indoor–outdoor metric depth benchmark in a VLM-compatible format. Experiments show that DepthVLM significantly outperforms existing VLMs with higher inference efficiency, surpasses leading pure vision models, and improves complex 3D spatial reasoning, moving toward a truly unified multimodal foundation model.
\end{abstract}

\section{Introduction}
\label{introduction}
With the rapid advancement of Large Language Models (LLMs)~\cite{chiang2023vicuna,liu2024deepseek,touvron2023llama,yang2025qwen3}, growing efforts have extended them beyond pure text understanding, giving rise to Vision-Language Models (VLMs)~\cite{jin2025streamingassistant,yu2026visiontrim,zhang2025videollama} that tackle diverse multimodal tasks. Despite strong performance on 2D tasks such as visual reasoning and image captioning, current VLMs remain limited in complex 3D understanding~\cite{chen2020scanrefer,majumdar2024openeqa,piccinelli2025unidepthv2,yang2025vsibench}, which is crucial for applications like AR/VR, autonomous driving, and embodied robotics.

\begin{figure}[t]
\centering
\includegraphics[width=1.0\linewidth]{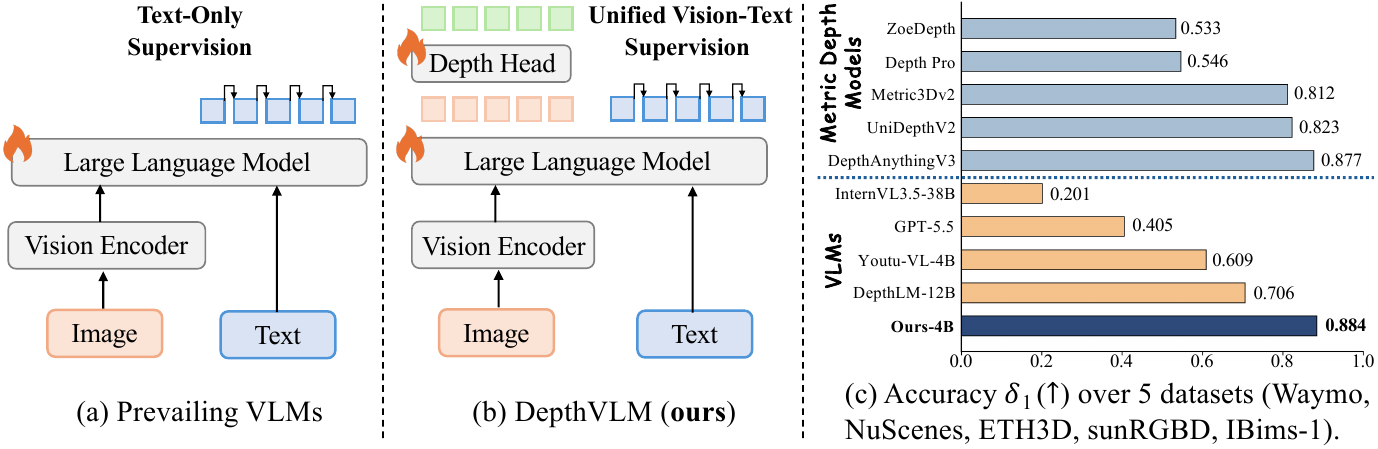}
\vspace{-4mm}
\caption{\textbf{Comparison of prevailing VLMs with our method.} (a) Prevailing VLMs are typically supervised solely in the text space, leaving dense 3D geometry out of reach. (b) DepthVLM introduces a unified vision–text supervision paradigm by integrating a lightweight depth head, natively enabling a single VLM backbone to generate dense geometry alongside language responses. (c) While even advanced VLMs such as GPT-5.5~\cite{singh2025gpt-5.4} struggle to infer 3D structure from 2D inputs, our model significantly outperforms prior VLMs and even surpasses leading specialized pure vision models.}
\label{teaser2}
\vspace{-5mm}
\end{figure}

A fundamental limitation of prevailing VLMs is their \emph{text-only supervision} paradigm: visual signals are consumed only as inputs, while outputs are generated as autoregressive text. This design inherently under-constrains fine-grained visual perception and prevents explicit modeling of dense scene geometry, as shown in Figure~\ref{teaser2}(a). To address this, prior works~\cite{fan2025vlm-3r,wu2025spatialmllm,zheng2025vgllm} inject geometric signals (\emph{e.g.}, depth maps or point clouds) from pretrained 3D models to augment VLMs, but such pipelines rely on knowledge distillation from external vision experts and inevitably suffer from error accumulation. More recent works~\cite{hu2025g2vlm,xu2025multi-spatialmllm,yan2026omnistream} instead explore direct geometric prediction from RGB inputs within VLMs. DepthLM~\cite{cai2025depthlm} first demonstrates that VLMs can match pure vision models on metric depth estimation, but its single-pixel query per inference makes dense prediction prohibitively slow, while its text-heavy supervision substantially degrades the VLM's general VQA capability. Youtu-VL~\cite{wei2026youtu-vl} further enables full-image depth prediction in one pass, yet its token-level outputs remain coarse and require post-hoc interpolation for pixel-level detail. Moreover, its from-scratch training recipe demands massive data and compute, limiting direct adaptation to existing VLMs.



These observations raise a natural question: \emph{can a VLM serve as a native dense geometry predictor with minimal architectural change, while preserving its general multimodal capability?} Focusing on dense metric depth estimation, a fundamental task in 3D understanding, we propose \textbf{DepthVLM}, a simple yet effective framework that enables a single VLM backbone to jointly generate dense pixel-level depth maps and language responses. As shown in Figure~\ref{teaser2}(b), we attach a lightweight depth head to the LLM backbone, taking processed visual tokens as input, and fine-tune the model under a \emph{unified vision–text supervision} paradigm. In a single forward pass, DepthVLM predicts full-image depth for all pixels without post-processing, reducing DepthLM's $\mathcal{O}(HW)$ inference cost to $\mathcal{O}(1)$. Moreover, unlike fixed-resolution vision models~\cite{wang2025vggt}, DepthVLM inherits the native-resolution flexibility of VLMs and can be seamlessly integrated into the standard instruction tuning stage.

Since extending VLMs to other tasks often degrades their general multimodal capability~\cite{dong2023dreamllm,zhang2024psalm}, we adopt a two-stage training strategy: Stage-1 trains only the added depth head to establish initial depth prediction ability, and Stage-2 fine-tunes the full model end-to-end. We further introduce \textbf{DepthVLM-Bench}, a unified benchmark that aggregates public indoor and outdoor depth datasets into a VLM-compatible format, enabling both effective training and fair comparison with pure vision models. Interestingly, we find that equipping VLMs with dense geometry prediction improves downstream 3D spatial reasoning performance, further highlighting the value of a unified foundation model that jointly excels at low-level dense geometry prediction and high-level multimodal understanding.

In summary, our contributions are threefold:
\begin{itemize}
    \item We find that a VLM can serve as a native dense geometry predictor and propose a lightweight recipe that yields a unified foundation model for both dense geometry generation and multimodal interaction, seamlessly compatible with the standard instruction-tuning stage.
    \item We devise a two-stage training strategy that preserves the VLM's original multimodal capability, and present DepthVLM-Bench, a unified indoor–outdoor benchmark that enables VLM training and direct comparison with pure vision models on metric depth estimation.
    \item Extensive experiments across diverse datasets show that DepthVLM significantly outperforms existing VLMs with higher inference efficiency, surpasses state-of-the-art pure vision models on metric depth estimation, and further improves 3D spatial reasoning performance.
\end{itemize}

\section{Related Work}
\label{related work}
\subsection{Dense Metric Depth Estimation}
Dense metric depth estimation aims to recover per-pixel absolute depth values from RGB images, which is fundamental for 3D scene understanding. Early methods~\cite{bhat2021adabins,eigen2014depth} rely on single-domain supervision, producing models specialized to either indoor rooms~\cite{nyuv2} or outdoor scenes~\cite{kitti} with limited cross-domain generalization. To improve robustness, MiDaS~\cite{ranftl2020midas} and DPT~\cite{ranftl2021dpt} introduce affine-invariant prediction across diverse datasets, but only provide relative depth without metric scale. To resolve scale ambiguity, ZoeDepth~\cite{bhat2023zoedepth} combines relative and metric depth via domain-specific heads, while Metric3D~\cite{yin2023metric3d,hu2024metric3dv2} unifies inputs in a canonical camera space. More recently, UniDepth~\cite{piccinelli2024unidepth,piccinelli2025unidepthv2} jointly estimates depth and camera intrinsics in a self-promptable manner, and DepthAnything~\cite{yang2024depthv1,yang2024depthv2,lin2025depthv3} leverages large-scale synthetic supervision for zero-shot generalization. Despite their strong geometric accuracy, these pure vision models focus solely on low-level geometric prediction and lack high-level language interaction, limiting their applicability to 3D reasoning tasks.


\subsection{VLMs for 3D Spatial Understanding}
\textbf{Spatial-Enhanced VLMs.}
To bridge the gap between 2D semantics and 3D spatial intelligence, a line of research augments VLMs~\cite{bai2025qwen3vl,zhang2024llava-video,wang2025internvl3.5} with external geometric signals. One direction~\cite{hong20233d,chen2024ll3da,zheng2025video3dllm,yu2025inst3d,zhu2024llava3d,huang2023leo,huang2024chat-scene,qi2025gpt4scene,wang2025n3d} directly feeds explicit 3D data (\emph{e.g.}, point clouds, voxels, or depth maps) from sensors into LLMs via projectors. While effective on 3D VQA benchmarks~\cite{azuma2022scanqa, ma2022sqa3d}, these methods rely on sparse and costly 3D data and are largely limited to indoor scenes. Another direction elicits spatial reasoning purely from 2D inputs. SpatialVLM~\cite{chen2024spatialvlm} and SpatialRGPT~\cite{cheng2024spatialrgpt} convert vision outputs into textual supervision, while Ross3D~\cite{wang2025ross3d} introduces multi-view reconstruction as an auxiliary objective. More recent works~\cite{wu2025spatialmllm,fan2025vlm-3r,zheng2025vgllm,huang20253drs,wu2026vega-3d} further distill geometric priors from 3D reconstruction~\cite{wang2025vggt,wang2025pi3,wang2025cut3r}  or video diffusion models~\cite{wan2025wan,blattmann2023stable} into VLMs to improve spatial reasoning. However, these methods rely on external vision experts, making them prone to error accumulation, and are still limited to textual outputs without enabling dense, pixel-level geometry prediction.


\textbf{Geometry-Generative VLMs.}
Recent studies~\cite{yan2026omnistream} instead treat the VLM as a unified foundation model that directly generates dense geometry from RGB inputs. Multi-SpatialMLLM~\cite{xu2025multi-spatialmllm} and Seed1.5-VL~\cite{guo2025seed1.5-vl} explore pixel-level metric depth estimation while lagging behind pure vision models. G$^2$VLM~\cite{hu2025g2vlm} adopts a Mixture-of-Experts architecture for unified modeling, yet focuses on relative depth. DepthLM~\cite{cai2025depthlm} matches advanced vision models in accuracy, but predicts only one pixel per inference and its text-heavy supervision severely degrades general performance. Youtu-VL~\cite{wei2026youtu-vl} enables full-image depth prediction in one pass, but produces coarse token-level outputs and relies on costly from-scratch training. In contrast, our method lightweightly equips existing VLMs with dense metric depth estimation while preserving their general capability. Inheriting native-resolution processing, it enables flexible inputs and can be seamlessly integrated into standard instruction tuning.

\section{Methodology}
\label{method}

\begin{figure}[t]
\centering
\includegraphics[width=1.0\linewidth]{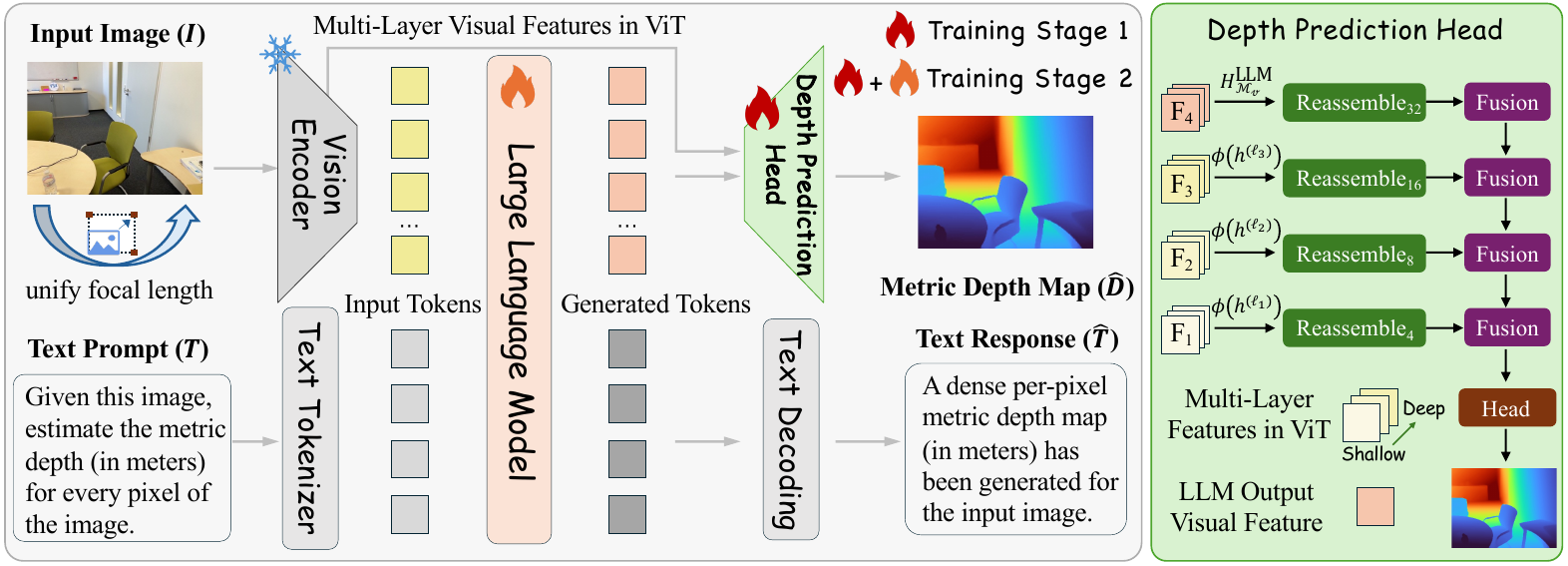}
\vspace{-4mm}
\caption{\textbf{Overview of our proposed DepthVLM.} We extend the standard VLM architecture with a lightweight DPT-style~\cite{ranftl2021dpt} depth prediction head, and adopt a two-stage training strategy to preserve the backbone's general VQA capability. In addition, input images are normalized to a unified focal length, eliminating camera-induced ambiguity across heterogeneous dataset domains.}
\label{pipeline}
\vspace{-5mm}
\end{figure}

Our goal is to develop a unified foundation model that natively supports both low-level dense geometry prediction and high-level multimodal understanding within a single VLM backbone. As illustrated in Figure~\ref{pipeline}, we (i) augment the standard VLM with a lightweight DPT-style~\cite{ranftl2021dpt} depth head to jointly produce dense metric depth map and language responses; (ii) employ a two-stage training strategy to preserve the VLM's inherent multimodal capability; and (iii) leverage a multi-source training corpus together with focal-length normalization to mitigate camera-induced ambiguity across heterogeneous sensors, yielding strong cross-dataset generalization.


\subsection{Model Architecture}
\label{sec:arch}

\textbf{Preliminaries.}
A standard VLM comprises three components: a vision encoder $\mathcal{E}_v$ that tokenizes an input image $I\!\in\!\mathbb{R}^{3\times H\times W}$ into $N_v$ vision tokens, a projector $\phi$ that maps them into the LLM embedding space, and an autoregressive language model $\mathcal{F}_{\text{LLM}}$ that processes the joint multimodal sequence to generate text. Given an image $I$ and a text prompt $T$, the VLM produces hidden states as
\begin{equation}
    H^{\text{LLM}} = \mathcal{F}_{\text{LLM}}\!\left([\,\phi(\mathcal{E}_v(I));\, T\,]\right) \in \mathbb{R}^{(N_v+N_t)\times d}.
\end{equation}

\textbf{Motivation: VLM as a Native Dense Predictor.}
Prior works on 2D dense understanding~\cite{wu2024visionllmv2} typically augment the VLM with region-level encoders~\cite{rasheed2024glamm} or task-specific tokens~\cite{tang2025ufo}, inevitably fragmenting the architecture and complicating the training and inference pipelines. Inspired by recent 3D foundation models~\cite{wang2025vggt, wang2025pi3} that derive dense geometry directly from transformer tokens, we instead ask: \emph{is a standard VLM already a dense predictor?} We answer this affirmatively by showing that dense geometry can be decoded directly from the VLM’s own vision tokens using a lightweight DPT-style~\cite{ranftl2021dpt} head over multi-scale visual features, without altering its text generation pathway.

\textbf{Unified Architecture for Dense Geometry.}
A key observation is that the vision encoder $\mathcal{E}_v$ naturally provides a hierarchy of representations---from low-level appearance cues in shallow layers to high-level semantics in deeper layers---that inherently form a multi-scale pyramid well suited for dense prediction. Let $\{h^{(\ell)}\}_{\ell=1}^{L_v}$ denote the per-layer hidden states of the ViT and $H^{\text{LLM}}$ the last-layer hidden states of the LLM. We extract four feature maps from the VLM: three intermediate ViT layers $\{\ell_1,\ell_2,\ell_3\}$ together with the LLM's final hidden states at image-token positions:
\begin{equation}
    F_k \;=\;
    \left\{
    \begin{array}{@{}l@{\quad}l@{\quad}l@{}}
        \phi\!\left(h^{(\ell_k)}\right) \in \mathbb{R}^{N_v\times d}, & k=1,2,3, & \text{(ViT intermediate layers)} \\[3pt]
        H^{\text{LLM}}_{\,\mathcal{M}_v} \in \mathbb{R}^{N_v\times d},  & k=4,     & \text{(LLM final layer)}
    \end{array}
    \right.
\end{equation}
where $\mathcal{M}_v$ selects LLM hidden states at image-token positions. $F_{1,2,3}$ capture purely visual features with increasing abstraction, while $F_4$ encodes vision-language contextualized representations.

Unlike the original DPT~\cite{ranftl2021dpt} that operates on native ViT features, visual tokens in a VLM are already downsampled by the patch merger~\cite{bai2025qwen3vl}. We therefore avoid additional downsampling and instead construct a bottom-up pyramid via upsampling, assigning higher spatial resolution to earlier ViT layers. Specifically, each ${F}_k$ is projected with a $1\!\times\!1$ convolution and resampled to a layer-specific resolution, yielding finer spatial details for shallower features. The resulting multi-scale features are fused with RefineNet blocks~\cite{lin2017refinenet} and decoded into a dense metric depth map at the input resolution:
\begin{equation}
    \hat{D} \;=\; \mathrm{DPT}\!\left(F_1, F_2, F_3, F_4\right) \in \mathbb{R}^{\,H\times W},
    \qquad
    \big(\hat{D},\, \hat{T}\big) \;=\; \mathrm{DepthVLM}(I, T),
\end{equation}
where a final $\mathrm{Softplus}$ activation ensures strictly positive depth values. In this way, our model jointly generates dense metric geometry $\hat{D}$ and text response $\hat{T}$ within a unified foundation model.

\subsection{Two-Stage Training Strategy}
\label{two-stage}
To introduce dense geometry prediction while preserving the original multimodal understanding, we adopt a two-stage training strategy. In the first stage, we train only the depth head to initialize dense depth prediction capability. In the second stage, we unfreeze the LLM backbone and fine-tune the model end-to-end, enabling tighter integration of geometric prediction with multimodal reasoning.

\textbf{Stage-1: Depth Head-Only Training.} Since the introduced depth head is randomly initialized, directly training it with the VLM can lead to noisy gradients that may disrupt pretrained knowledge. We therefore freeze the entire VLM and train only the depth head. Following standard practice~\cite{hu2024metric3dv2,yang2024depthv2}, we supervise the predicted depth map $\hat{D}$ using the scale-invariant logarithmic (SILog) loss~\cite{eigen2014depth}:
\begin{equation}
    \mathcal{L}_{\mathrm{depth}} \;=\; \sqrt{\frac{1}{|\Omega|}\sum_{i\in\Omega} d_i^{\,2} \;-\; \lambda\Big(\frac{1}{|\Omega|}\sum_{i\in\Omega} d_i\Big)^{\!2}},
    \qquad d_i = \log \hat{D}_i - \log D_i^{*},
\end{equation}
where $\Omega$ denotes pixels with valid ground-truth depth $D^{*}$ and $\lambda$ provides a balanced inductive bias, preserving metric supervision while reducing sensitivity to dataset-specific scale variations.


\textbf{Stage-2: End-to-End Fine-Tuning.}
To further strengthen geometric prediction in synergy with the VLM's inherent language interaction capability, we unfreeze the LLM backbone and perform end-to-end fine-tuning on a mixture of instruction-following data. The overall objective is a weighted combination of the autoregressive language modeling loss and the depth loss defined in Stage-1:
\begin{equation}
    \mathcal{L}_{\mathrm{joint}} \;=\; \mathcal{L}_{\mathrm{text}} \;+\; \alpha\,\mathcal{L}_{\mathrm{depth}},
    \qquad
    \mathcal{L}_{\mathrm{text}} \;=\; -\sum_{t} \log p_{\theta}\!\left(\hat{T}_t \,\big|\, \hat{T}_{<t},\, I,\, T\right),
\end{equation}
where $\mathcal{L}_{\mathrm{text}}$ is the standard cross-entropy loss over response tokens and $\alpha$ balances the two objectives. 

\subsection{Mixed-Source Data Curation}
\label{mixed-data}
\textbf{Eliminating Camera Ambiguity.} Joint training across datasets suffers from camera-induced scale ambiguity in metric depth estimation. Images with different focal lengths can depict similar scenes but correspond to inconsistent metric depths, leading to conflicting supervision and poor generalization. We address this by adopting focal-length normalization following prior works~\cite{cai2025depthlm,piccinelli2024unidepth}, rescaling all images to a unified focal length $f_c$ to remove dataset-specific biases and enforce consistent pixel-to-metric mapping. Formally, given an image $I$ with focal length $f$ and depth map $D$, we apply:
\begin{equation}
    s = {f_c} \;/\; {f}, \qquad \tilde{I} \;=\; \mathcal{R}_{s}(I),
    \qquad
    \tilde{D} \;=\; \mathcal{R}_{s}(D),
\end{equation}
where $\mathcal{R}_{s}(\cdot)$ denotes isotropic bilinear resizing. After normalization, all samples are aligned to a virtual camera with focal length $f_c$. This removes cross-dataset scale discrepancies and enables the model to learn a focal-invariant mapping that generalizes well to open-world images.


\textbf{DepthVLM-Bench.} We assemble a diverse set of widely used public datasets for metric depth estimation into a unified benchmark that supports training VLMs for dense geometry prediction and enables direct comparison with pure vision models under a consistent protocol.


\textit{Training split.} We mix the training set of 8 datasets covering indoor and outdoor scenes. For indoor data, we use ScanNet++~\cite{yeshwanth2023scannet++}, Taskonomy~\cite{zamir2018taskonomy}, HM3D~\cite{ramakrishnan2021hm3d}, and Matterport3D~\cite{chang2017matterport3d}; for outdoor data, we use Argoverse2~\cite{wilson2023argoverse}, Waymo~\cite{sun2020scalability}, DDAD~\cite{guizilini2020ddad}, and NuScenes~\cite{caesar2020nuScenes}. In contrast to pure vision models~\cite{bochkovskii2024depthpro,lin2025depthv3}, which often rely on more than 20 datasets with extensive synthetic data, our model achieves comparable performance with an order of magnitude less data.

\textit{Evaluation split.} We evaluate on 9 datasets across domains, all disjoint from the training set: 4 indoor (ScanNet++, sunRGBD~\cite{song2015sun}, IBims-1~\cite{koch2018ibims1}, NYUv2~\cite{nyuv2}), 4 outdoor (Argoverse2, Waymo, DDAD, NuScenes), and ETH3D~\cite{schops2017eth3d} containing both indoor and outdoor scenes. For each dataset, we sample 1k images and 10 pixels per image (10k pixels total), oversampling smaller datasets when needed.

\definecolor{OutBlue}{HTML}{DCE6F7}     
\definecolor{MixPurple}{HTML}{E6DCF0}   
\definecolor{InPink}{HTML}{F7DCDC}      
\definecolor{SectionGray}{HTML}{ECECEC} 
\definecolor{mygray}{gray}{0.6}
\definecolor{mygreen}{RGB}{0, 100, 0}
\definecolor{myred}{RGB}{255, 0, 0}

\newcolumntype{C}{>{\centering\arraybackslash}X}

\begin{table}[t]
\centering
\caption{\textbf{Comparison with existing VLMs on metric depth estimation across diverse indoor and outdoor datasets.} For VLMs not explicitly trained for this task, we adopt the prompting strategy proposed in DepthLM~\cite{cai2025depthlm} to elicit their best performance. Even the state-of-the-art GPT-5.5~\cite{singh2025gpt-5.4} attains a $\delta_1$ of only around 0.4, highlighting the difficulty of the task for prevailing VLMs. \textbf{Bold} and \underline{underlined} values denote the best and second-best results, respectively. }
\label{tab:delta1_vlms}
\scriptsize
\setlength{\tabcolsep}{1.5pt}
\renewcommand{\arraystretch}{1.15}
\begin{tabularx}{\linewidth}{l *{4}{C} C *{4}{C} C}
\toprule
\multirow{2}{*}{\makecell[l]{$\delta_1(\uparrow)$ of various methods}}
  & \multicolumn{4}{c}{\cellcolor{OutBlue}\textit{Outdoor}}
  & \cellcolor{MixPurple}\textit{Out+In}
  & \multicolumn{4}{c}{\cellcolor{InPink}\textit{Indoor}}
  & \multirow{2}{*}{\textbf{Avg.}} \\
  & Argoverse2 & Waymo & DDAD & NuScenes
  & ETH3D
  & ScanNet++ & sunRGBD & IBims-1 & NYUv2 & \\
\midrule

\rowcolor{SectionGray}
\multicolumn{11}{c}{\textit{Naive Prediction with Constant Answers}} \\
Always Output 2.0m & 0.002 & 0.004 & 0.002 & 0.010 & 0.112 & 0.261 & 0.380 & 0.269 & 0.373 & 0.157 \\

\rowcolor{SectionGray}
\multicolumn{11}{c}{\textit{General-Purpose VLMs}} \\
GPT-4o~\cite{hurst2024gpt-4o}       & 0.141 & 0.139 & 0.193 & 0.174 & 0.296 & 0.369 & 0.358 & 0.393 & 0.394 & 0.273 \\
GPT-5.5~\cite{singh2025gpt-5.4}      & 0.378 & 0.368 & 0.304 & 0.276 & 0.369 & 0.432 & 0.527 & 0.483 & 0.525 & 0.407  \\
Qwen3-VL-4B~\cite{bai2025qwen3vl}   & 0.033 & 0.006 & 0.086 & 0.040 & 0.085 & 0.208 & 0.220 & 0.037 & 0.155 & 0.097 \\
Qwen3-VL-8B~\cite{bai2025qwen3vl}   & 0.119 & 0.061 & 0.243 & 0.118 & 0.175 & 0.274 & 0.286 & 0.081 & 0.194 & 0.172 \\ 
Qwen3-VL-32B~\cite{bai2025qwen3vl}  & 0.029 & 0.017 & 0.138 & 0.049 & 0.150 & 0.434 & 0.536 & 0.105 & 0.435 & 0.210 \\
InternVL3.5-8B~\cite{wang2025internvl3.5} & 0.139 & 0.110 & 0.177 & 0.111 & 0.185 & 0.395 & 0.459 & 0.214 & 0.431 & 0.247 \\
InternVL3.5-14B~\cite{wang2025internvl3.5}  & 0.086 & 0.045 & 0.128 & 0.084 & 0.183 & 0.390 & 0.440 & 0.235 & 0.445 & 0.226 \\
InternVL3.5-38B~\cite{wang2025internvl3.5}  & 0.131 & 0.129 & 0.219 & 0.110 & 0.181 & 0.400 & 0.431 & 0.155 & 0.423 & 0.242 \\

\rowcolor{SectionGray}
\multicolumn{11}{c}{\textit{Spatial-Enhanced VLMs} }\\
SpaceLLaVA-13B~\cite{chen2024spatialvlm}  & 0.006 & 0.002 & 0.001 & 0.006 & 0.107 & 0.050 & 0.044 & 0.172 & 0.087 & 0.053 \\
SpatialRGPT-8B~\cite{cheng2024spatialrgpt}  & 0.045 & 0.064 & 0.096 & 0.116 & 0.133 & 0.124 & 0.084 & 0.044 & 0.070 & 0.086  \\
Cambrian-S-7B~\cite{yang2025cambrians}  & 0.006 & 0.019 & 0.038 & 0.033 & 0.069 & 0.145 & 0.073 & 0.057 & 0.063 & 0.056  \\
\rowcolor{SectionGray}
\multicolumn{11}{c}{\textit{VLMs Trained on Metric Depth Estimation}} \\
Youtu-VL-4B~\cite{wei2026youtu-vl} & 0.663 & 0.473 & 0.342 & 0.698 & 0.286 & 0.522 & 0.734 & 0.856 & 0.849 & 0.603 \\
DepthLM-12B~\cite{cai2025depthlm}  & 0.761 & 0.588 & 0.654 & 0.736 & 0.666 & 0.756 & 0.785 & 0.754 & 0.866 & 0.730 \\

\textbf{Ours-4B}  & \textbf{0.810} & \textbf{0.879} & \textbf{0.818} & \underline{0.821} & \underline{0.924} & \underline{0.861} & \underline{0.882} & \underline{0.912} & \underline{0.908} & \underline{0.868} \\
\textbf{Ours-8B}  & \underline{0.798} & \underline{0.865} & \underline{0.813} & \textbf{0.831} & \textbf{0.928} & \textbf{0.901} & \textbf{0.889} & \textbf{0.936} & \textbf{0.920} & \textbf{0.876} \\

\bottomrule
\end{tabularx}
\vspace{-5mm}
\end{table}

\newcolumntype{C}{>{\centering\arraybackslash}X}
\begin{table}[t]
\centering
\caption{\textbf{Comparison with specialized pure vision models on metric depth estimation across indoor and outdoor datasets.} Despite being a unified VLM that preserves strong multimodal capabilities, our method can outperform state-of-the-art pure vision specialists, demonstrating that dense geometry prediction can emerge natively within a single vision-language foundation model.}
\label{tab:delta1_purevision}
\footnotesize
\setlength{\tabcolsep}{2pt}
\renewcommand{\arraystretch}{1.15}
\begin{tabularx}{\linewidth}{l *{2}{C} C *{2}{C} c}
\toprule
\multirow{2}{*}{\makecell[l]{$\delta_1(\uparrow)$ of various methods}}
  & \multicolumn{2}{c}{\cellcolor{OutBlue}\textit{Outdoor}}
  & \cellcolor{MixPurple}\textit{Out+In}
  & \multicolumn{2}{c}{\cellcolor{InPink}\textit{Indoor}}
  & \multirow{2}{*}{\textbf{Avg.}} \\
  & Waymo & NuScenes
  & ETH3D
  & sunRGBD & IBims-1
  & \\
\midrule
ZoeDepth~\cite{bhat2023zoedepth}            & 0.639 & 0.196 & 0.345 & 0.769 & 0.718 & 0.533 \\
Depth Pro~\cite{bochkovskii2024depthpro}    & 0.255 & 0.389 & 0.355 & 0.852 & 0.880 & 0.546 \\
Metric3D~\cite{yin2023metric3d}             & 0.879 & 0.721 & 0.373 & 0.222 & 0.796 & 0.598 \\
Metric3Dv2~\cite{hu2024metric3dv2}          & {0.923} & 0.747 & 0.851 & 0.813 & 0.726 & 0.812 \\
UniDepth~\cite{piccinelli2024unidepth}      & 0.670 & {0.858} & 0.149 & 0.907 & 0.158 & 0.548 \\
UniDepthV2~\cite{piccinelli2025unidepthv2}  & 0.730 & {0.872} & 0.657 & {0.911} & {0.941} & 0.823 \\
DepthAnything~\cite{yang2024depthv1}        & 0.739 & 0.205 & 0.277 & 0.847 & 0.854 & 0.584 \\
DepthAnythingV2~\cite{yang2024depthv2}      & 0.715 & 0.168 & 0.111 & 0.697 & 0.887 & 0.516 \\
DepthAnythingV3~\cite{lin2025depthv3}       & {0.885} & 0.790 & 0.843 & {0.913} & {0.955} & 0.877 \\
\midrule
\textbf{Ours-4B} & 0.879 & 0.821 & {0.924} & 0.882 & 0.912 & \underline{0.884} \\
\textbf{Ours-8B} & 0.865 & 0.831 & {0.928} & 0.889 & 0.936  & \textbf{0.890} \\
\bottomrule
\end{tabularx}
\vspace{-5mm}
\end{table}

\section{Experiment}
\label{experiments}
\subsection{Experimental Settings}
\textbf{Baselines and Metrics.} We compare our model against VLMs and pure vision models. Baselines include four groups: (i) \emph{general-purpose VLMs}: Qwen3-VL~\cite{bai2025qwen3vl}, InternVL3.5~\cite{wang2025internvl3.5}, GPT-4o~\cite{hurst2024gpt-4o}, GPT-5.5~\cite{singh2025gpt-5.4}; (ii) \emph{spatially-enhanced VLMs}: SpaceLLaVA-13B~\cite{chen2024spatialvlm}, SpatialRGPT-8B~\cite{cheng2024spatialrgpt}, Cambrian-S-7B~\cite{yang2025cambrians}; (iii) \emph{depth-specialized VLMs}: Youtu-VL-4B~\cite{wei2026youtu-vl}, DepthLM-12B~\cite{cai2025depthlm}; and (iv) \emph{pure vision models}: ZoeDepth~\cite{bhat2023zoedepth}, Depth Pro~\cite{bochkovskii2024depthpro}, UniDepth~\cite{piccinelli2024unidepth,piccinelli2025unidepthv2}, Metric3D~\cite{yin2023metric3d,hu2024metric3dv2}, DepthAnything~\cite{yang2024depthv1,yang2024depthv2,lin2025depthv3}. Following standard practice, we report $\delta_1$ accuracy, the percentage of predictions within $25\%$ relative error of ground truth. All models are evaluated on the DepthVLM-Bench evaluation split.

\textbf{Implementation Details.}
We adopt Qwen3-VL~\cite{bai2025qwen3vl} (4B/8B) as the default VLM backbone, and integrate a lightweight DPT-style~\cite{ranftl2021dpt} head with $34$M parameters ($<\!1\%$ of the LLM). Models are trained in PyTorch on $4.4$M samples from the training split of DepthVLM-Bench with uniform sampling. Intermediate ViT features are taken from layers 5, 11, and 17 for 4B, and 8, 16, and 24 for 8B. We use AdamW with a cosine schedule, learning rates of $3.5\!\times\!10^{-4}$ and $2\!\times\!10^{-5}$, and warmup ratios of $0.04$ and $0.05$ for Stage-1 and Stage-2. The balance factors $\lambda$ and $\alpha$ are set to $0.5$ and $1.0$.

\begin{table}[t]
\centering
\begin{threeparttable}
\caption{\textbf{Evaluation on broad visual benchmarks, covering general VQA, document understanding, multi-image reasoning, counting, and hallucination.} Empowered by our lightweight depth head and two-stage training strategy, our method natively gains the ability to generate dense geometry \emph{without sacrificing} the general multimodal capability of the underlying VLM, in sharp contrast to prior text-heavy supervision approach~\cite{cai2025depthlm} that typically incurs substantial capability degradation.}
\label{tab:general_VQA}
\scriptsize
\setlength{\tabcolsep}{1pt}
\renewcommand{\arraystretch}{1.15}
\begin{tabularx}{\linewidth}{l *{8}{C}}
\toprule
\textbf{Methods} & \textbf{MMB-EN} & \textbf{MMB-CN} & \textbf{MMStar} & \textbf{ScienceQA} & \textbf{BLINK} & \textbf{OCRBench} & \textbf{CountBench} & \textbf{POPE} \\
\midrule
\textcolor{mygray}{Pixtral-12B~\cite{agrawal2024pixtral}} & \textcolor{mygray}{78.2} & \textcolor{mygray}{73.7} & \textcolor{mygray}{52.0} & \textcolor{mygray}{88.8} & \textcolor{mygray}{49.2} & \textcolor{mygray}{660} & \textcolor{mygray}{69.5} & \textcolor{mygray}{85.5} \\
\textcolor{mygray}{Qwen3-VL-4B~\cite{bai2025qwen3vl}} & \textcolor{mygray}{83.4} & \textcolor{mygray}{81.1} & \textcolor{mygray}{60.9} & \textcolor{mygray}{91.2} & \textcolor{mygray}{63.8} & \textcolor{mygray}{817} & \textcolor{mygray}{97.7} & \textcolor{mygray}{89.8} \\
\textcolor{mygray}{Qwen3-VL-8B~\cite{bai2025qwen3vl}} & \textcolor{mygray}{84.7} & \textcolor{mygray}{82.4} & \textcolor{mygray}{63.4} & \textcolor{mygray}{92.8} & \textcolor{mygray}{65.0} & \textcolor{mygray}{833} & \textcolor{mygray}{98.3} & \textcolor{mygray}{88.8} \\
\midrule
DepthLM-12B~\cite{cai2025depthlm}$^\dagger$ & N/A & N/A & N/A & N/A & N/A & N/A & N/A & N/A \\
\rowcolor{gray!15}
\textbf{Ours-4B}
& 82.9 \textcolor{mygreen}{($\downarrow$ 0.5)}
& 81.8 \textcolor{myred}{($\uparrow$ 0.7)}
& 60.4 \textcolor{mygreen}{($\downarrow$ 0.5)}
& 91.3 \textcolor{myred}{($\uparrow$ 0.1)}
& 63.3 \textcolor{mygreen}{($\downarrow$ 0.5)}
& 832 \textcolor{myred}{($\uparrow$ 15)}
& 98.0 \textcolor{myred}{($\uparrow$ 0.3)}
& 89.9 \textcolor{myred}{($\uparrow$ 0.1)} \\ 
\rowcolor{gray!15}
\textbf{Ours-8B}
& 84.6 \textcolor{mygreen}{($\downarrow$ 0.1)}
& 82.3 \textcolor{mygreen}{($\downarrow$ 0.1)}
& 63.8 \textcolor{myred}{($\uparrow$ 0.4)}
& 93.1 \textcolor{myred}{($\uparrow$ 0.3)}
& 64.8 \textcolor{mygreen}{($\downarrow$ 0.2)}
& 862 \textcolor{myred}{($\uparrow$ 29)}
& 98.2 \textcolor{mygreen}{($\downarrow$ 0.1)}
& 89.1 \textcolor{myred}{($\uparrow$ 0.3)} \\
\bottomrule
\end{tabularx}
\begin{tablenotes}
\footnotesize
\item[$\dagger$] Due to its text-dominant supervised fine-tuning on single-pixel depth query, DepthLM~\cite{cai2025depthlm} collapses to always emitting a depth value regardless of the input instruction, making it incompatible with standard VQA evaluation protocols.
\end{tablenotes}
\end{threeparttable}
\vspace{-5mm}
\end{table} 

\begin{figure}[h]
\centering
\includegraphics[width=1.0\linewidth]{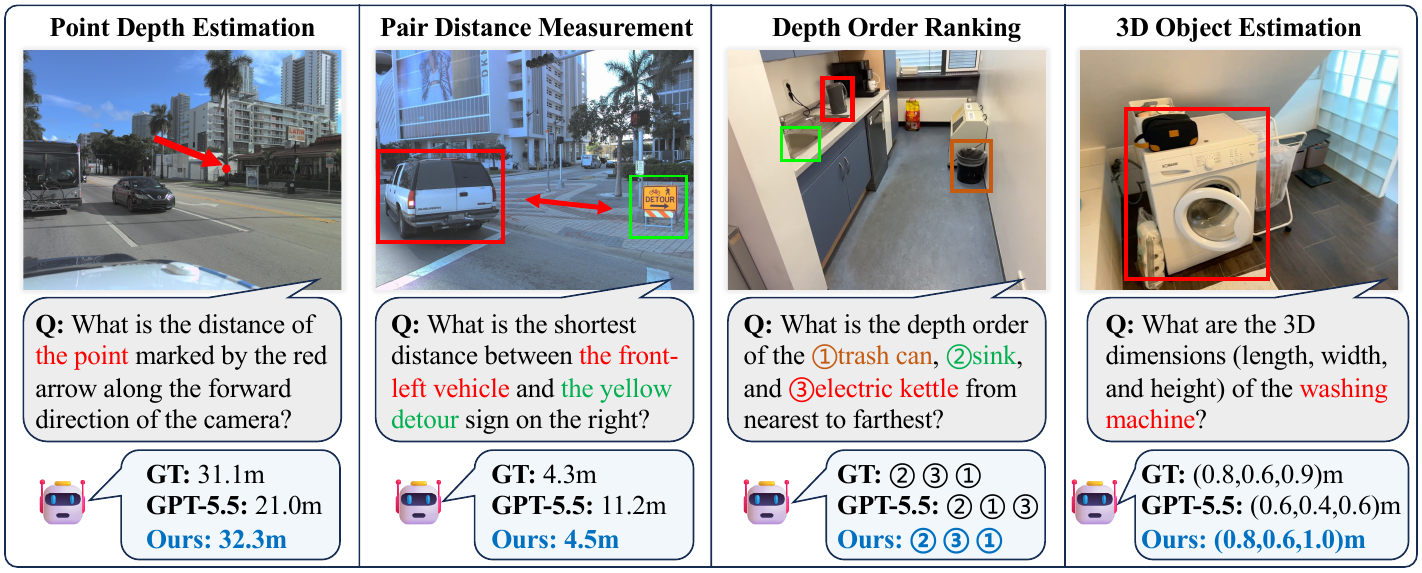}
\vspace{-4mm}
\caption{\textbf{Qualitative results on more complex 3D tasks.} Beyond dense metric depth estimation, our model further supports a variety of downstream 3D reasoning tasks, demonstrating that native dense geometry prediction serves as a solid foundation for high-level spatial reasoning in VLMs.}
\label{demo}
\vspace{-5mm}
\end{figure}
\subsection{Main Results}
\textbf{Comparison with Other VLMs.}
To evaluate metric depth estimation in existing VLMs, we follow DepthLM~\cite{cai2025depthlm} by prompting models with an arrow-marked pixel to predict its depth. As shown in Table~\ref{tab:delta1_vlms}, general-purpose VLMs perform poorly—especially in outdoor driving scenes—with Qwen3-VL-32B~\cite{bai2025qwen3vl} achieving $\delta_1=0.21$ and GPT-5.5~\cite{singh2025gpt-5.4} only $0.41$ on average, revealing a substantial gap to reliable 3D understanding. Even spatially enhanced VLMs, despite depth and calibration supervision, underperform a constant-depth baseline. In contrast, our model consistently excels across indoor and outdoor settings, significantly outperforming both larger and task-specific VLMs.

\textbf{Comparison with Pure Vision Models.}
Table~\ref{tab:delta1_purevision} further compares our model with leading specialized pure vision models on indoor and outdoor metric depth estimation. Since both pure vision models and DepthVLM produce dense metric depth maps, we evaluate them on the same sampled pixels used in the VLM setting for a fair comparison. Despite being a unified model with strong multimodal capabilities, our method not only significantly outperforms most vision specialists, including UniDepthV2~\cite{piccinelli2025unidepthv2} and Metric3Dv2~\cite{hu2024metric3dv2}, but also surpasses the state-of-the-art DepthAnythingV3~\cite{lin2025depthv3}.

\textbf{Evaluation on General Visual Benchmarks.}
To verify that dense geometry prediction does not compromise multimodal understanding, we evaluate on broad visual benchmarks in Table~\ref{tab:general_VQA}. Our models match their original VLM backbones and even improve on OCRBench~\cite{liu2023ocrbench} and POPE~\cite{li2023pope}. In contrast, prior depth-specialized VLMs such as DepthLM~\cite{cai2025depthlm} often overfit to text-heavy supervision and lose general-purpose capabilities. These results underscore the effectiveness of our unified design, which supports both accurate dense geometry prediction and strong multimodal understanding.

\textbf{Evaluation on Spatial Reasoning Tasks.}
We further find that enabling a VLM to act as a native 3D dense geometry predictor also improves spatial reasoning performance. Figure~\ref{demo} demonstrates more complex 3D reasoning tasks beyond metric depth estimation, where even pioneering GPT-5.5~\cite{singh2025gpt-5.4} may fail. These results suggest that strong native dense geometry prediction capabilities provide a solid foundation for high-level spatial reasoning in VLMs.

\begin{figure}[t]
\centering
\includegraphics[width=1.0\linewidth]{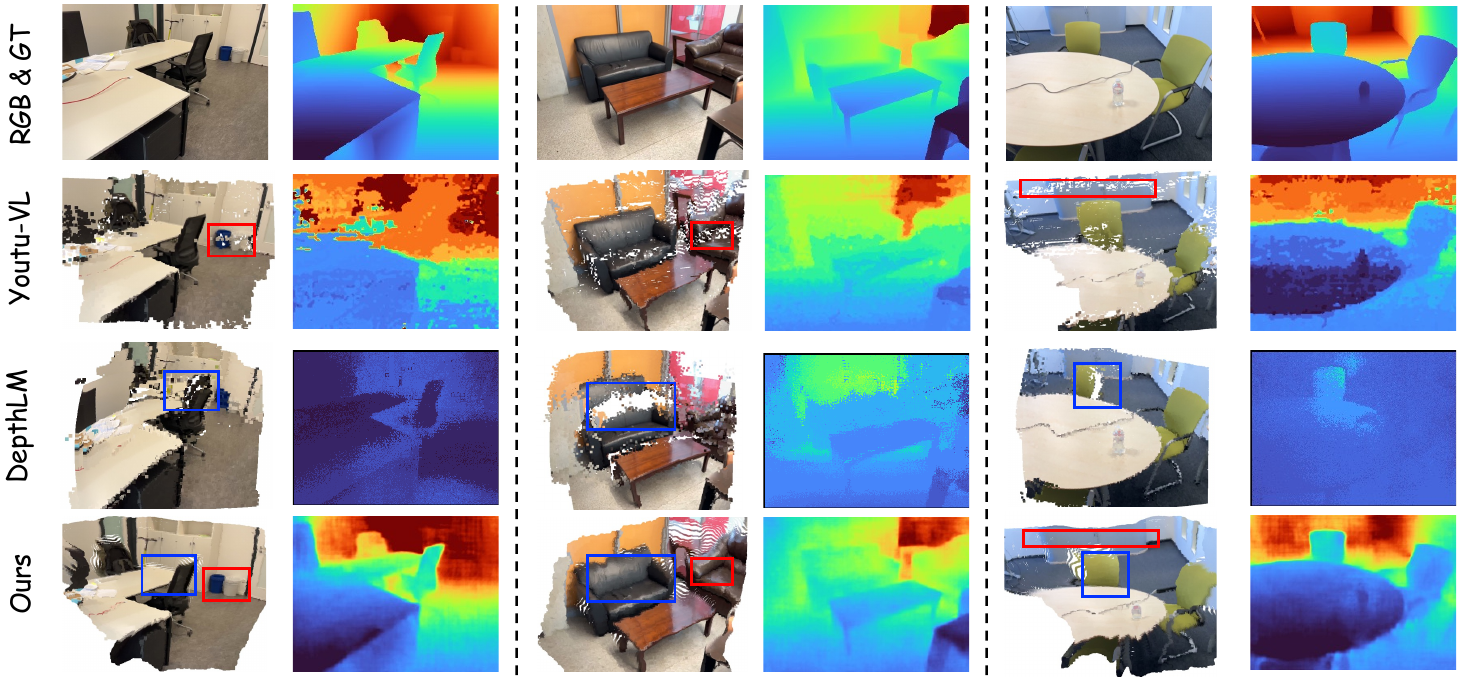}
\vspace{-4mm}
\caption{\textbf{Qualitative comparison with others.} Our results show finer structural details and improved semantic consistency across diverse scenes. Depth is color-coded from near (\spectral) to far.}
\label{visualization}
\vspace{-5mm}
\end{figure}


\begin{table}[t]
\centering
\begin{minipage}[t]{0.52\linewidth}
\centering
\captionof{table}{\textbf{Ablation of depth head designs.} ``Multi-scale'' indicates aggregation of visual features from intermediate ViT layers. Our lightweight DPT-style~\cite{ranftl2021dpt} head with multi-scale fusion performs best due to the design tailored to VLM features.}
\label{tab:ablation_depth_head}
\setlength{\tabcolsep}{3pt}
\renewcommand{\arraystretch}{1.15}
\resizebox{\linewidth}{!}{%
\begin{tabular}{l|c|cccc}
\toprule
\textbf{Depth Head} & \textbf{Multi-scale} & \textbf{Waymo} & \textbf{NuScenes} & \textbf{sunRGBD} & \textbf{IBims-1} \\
\midrule
Two-layer MLP            & \xmark & 0.547 & 0.444 & 0.533 & 0.695 \\
Two-layer MLP            & \cmark  & 0.723 & 0.776 & 0.727 & 0.806 \\
Original DPT~\cite{ranftl2021dpt} & \cmark & 0.866 & \textbf{0.826} & 0.856 & 0.895 \\
\rowcolor{gray!15}
\textbf{Ours (Lightweight DPT)} & \cmark & \textbf{0.879} & 0.821 & \textbf{0.882} & \textbf{0.912} \\
\bottomrule
\end{tabular}%
}
\end{minipage}
\vspace{-3mm}
\hfill
\begin{minipage}[t]{0.46\linewidth}
\centering
\captionof{table}{\textbf{Ablation of feature sources for the depth head.} ``Inter.'' denotes three intermediate layers and ``Final'' the last-layer output. Fusing multi-scale ViT features with the LLM final feature yields the best depth fidelity.}
\label{tab:ablation_feature_source}
\setlength{\tabcolsep}{3pt}
\renewcommand{\arraystretch}{1.34}
\resizebox{\linewidth}{!}{%
\begin{tabular}{l|cccc}
\toprule
\textbf{Feature Source} & \textbf{Waymo} & \textbf{NuScenes} & \textbf{sunRGBD} & \textbf{IBims-1} \\
\midrule
ViT Inter. + ViT Final          & 0.758 & 0.712 & 0.798 & 0.777 \\
LLM Inter. + LLM Final (Stage-1 only)   & 0.720 & 0.710 & 0.763 & 0.744 \\
LLM Inter. + LLM Final (Two-stage)          & 0.812 & 0.770 & 0.846 & 0.829 \\
\rowcolor{gray!15}
\textbf{ViT Inter. + LLM Final (Two-stage)} & \textbf{0.879} & \textbf{0.821} & \textbf{0.882} & \textbf{0.912} \\
\bottomrule
\end{tabular}%
}
\end{minipage}
\vspace{-3mm}
\end{table}

\textbf{Qualitative Visualizations.} As shown in Figure~\ref{visualization}, we compare the depth maps and corresponding 3D point clouds generated by Youtu-VL-4B~\cite{wei2026youtu-vl} and DepthLM-12B~\cite{cai2025depthlm} across diverse scenes. Youtu-VL produces noisy and fragmented point clouds with poor geometric continuity, while DepthLM maintains better semantic coherence but loses fine structural details. In contrast, our method significantly improves generation quality, preserving both semantic consistency and detailed spatial structure.

\begin{table}[t]
\centering
\caption{\textbf{Ablation of training strategies.} We compare four variants: (i) \textit{Stage-1 Only}, training only the depth head with the VLM frozen; (ii) \textit{Stage-2 Only}, directly fine-tuning the full model; (iii) \textit{Stage-1 + Stage-2$^\ddagger$}, where the vision encoder is unfrozen in Stage-2; and (iv) our full strategy. Unfreezing the vision encoder yields marginal depth gains but degrades general multimodal performance, whereas our design achieves strong depth estimation accuracy while preserving the VLM's general capability.}
\label{tab:training_strategy}
\renewcommand{\arraystretch}{1.25}
\setlength{\tabcolsep}{4pt}
\resizebox{\textwidth}{!}{%
\begin{tabular}{l|cccc|cccc}
\toprule
\multirow{2}{*}{\textbf{Training Strategy}} 
& \multicolumn{4}{c|}{\textbf{Depth Estimation} ($\delta_1\uparrow$)} 
& \multicolumn{4}{c}{\textbf{General Visual Benchmarks}} \\
\cmidrule(lr){2-5} \cmidrule(lr){6-9}
& \textbf{Waymo} & \textbf{NuScenes} & \textbf{sunRGBD} & \textbf{IBims-1} 
& \textbf{MMB-EN} & \textbf{MMStar} & \textbf{BLINK} & \textbf{OCRBench} \\
\midrule
Stage-1 Only   & 0.737 & 0.742 & 0.782 & 0.753 & \textbf{83.23} & \underline{60.36} & \textbf{63.55} & \textbf{840} \\
Stage-2 Only   & 0.784 & 0.762 & 0.826 & 0.805 & 81.44 & 57.21 & 57.77 & 793 \\
Stage-1 + Stage-2$^\ddagger$ (unfreeze ViT) & \textbf{0.884} & \textbf{0.837} & \textbf{0.893} & \underline{0.900} & 82.13 & 54.60 & 59.47 & 769 \\
\rowcolor{gray!15}
\textbf{Stage-1 + Stage-2 (freeze ViT)}  & \underline{0.879} & \underline{0.821} & \underline{0.882} & \textbf{0.912} & \underline{82.93} & \textbf{60.42} & \underline{63.25} & \underline{832} \\ 
\bottomrule
\end{tabular}%
}
\vspace{-5mm}
\end{table}

\begin{table}[t]
\centering
\begin{minipage}[t]{0.49\linewidth}
\centering
\captionof{table}{\textbf{Ablation of focal-length normalization.} We compare training on raw mixed-source images (\textit{w/o normalization}) with canonicalizing inputs to a shared focal length $f_{\mathrm{c}}\!\in\!\{800,\,1000,\,1200\}$. Normalization consistently improves performance, with $f_{\mathrm{c}}\!=\!1000$ achieving the best results across diverse benchmarks.}
\label{tab:ablation_focal_length}
\setlength{\tabcolsep}{4pt}
\renewcommand{\arraystretch}{1.09}
\resizebox{\linewidth}{!}{%
\begin{tabular}{l|cccc}
\toprule
\textbf{Training Setting} & \textbf{Waymo} & \textbf{NuScenes} & \textbf{sunRGBD} & \textbf{IBims-1} \\
\midrule
Raw mixed-source & 0.802 & 0.715 & 0.770 & 0.630 \\
Canonical $f_{\mathrm{c}}\!=\!800$  & 0.833 & 0.824 & 0.865 & 0.883 \\
\rowcolor{gray!15}
Canonical $f_{\mathrm{c}}\!=\!1000$ & \textbf{0.879} & 0.821 & \textbf{0.882} & \textbf{0.912} \\
Canonical $f_{\mathrm{c}}\!=\!1200$ & 0.858 & \textbf{0.840} & 0.837 & 0.856 \\
\bottomrule
\end{tabular}%
}
\end{minipage}
\vspace{-3mm}
\hfill
\begin{minipage}[t]{0.49\linewidth}
\centering
\captionof{table}{\textbf{Efficiency comparison with others.} We report the end-to-end cost of generating a $256{\times}192$ depth map. DepthLM~\cite{cai2025depthlm} uses per-pixel queries, while Youtu-VL~\cite{wei2026youtu-vl} predicts a sparse grid with upsampling. In contrast, our model outputs a pixel-aligned depth map in one pass, achieving higher efficiency without post-processing.}
\label{tab:efficiency}
\setlength{\tabcolsep}{2pt}
\renewcommand{\arraystretch}{1.2}
\resizebox{\linewidth}{!}{%
\begin{tabular}{l|c|c|c|c}
\toprule
\textbf{Method} &
\makecell{\textbf{\#Fwd.} \\ \textbf{/ image}} &
\makecell{\textbf{Output} \\ \textbf{Pattern}} &
\makecell{\textbf{Post-} \\ \textbf{proc.}} &
\makecell{\textbf{Latency} \\ \textbf{(ms)}} $\downarrow$ \\
\midrule
DepthLM-12B~\cite{cai2025depthlm}
 & $H{\times}W$ & point-wise queries & none
 & 13h \\
Youtu-VL-4B~\cite{wei2026youtu-vl}
 & 1 & sparse patch-level & bilinear $\uparrow$
 & 2.48s \\
\rowcolor{gray!15}
\textbf{Ours-4B}
 & {1} & {dense pixel-level} & {none}
 & \textbf{0.42s} \\
\bottomrule
\end{tabular}%
}
\end{minipage}
\vspace{-3mm}
\end{table}

\subsection{Ablation Studies}
In this section, we perform ablation experiments on the 4B model to thoroughly evaluate the effectiveness of the components.

\textbf{Ablation of Depth Head Variants.}
We compare different depth prediction heads in Table~\ref{tab:ablation_depth_head}. A two-layer MLP performs worst due to its overly simple architecture, while incorporating multi-scale ViT features already brings clear improvements. The original DPT~\cite{ranftl2021dpt} head remains suboptimal because downsampling the LLM final visual feature discards high-level semantic information. In contrast, our lightweight DPT-style head constructs a bottom-up feature pyramid through upsampling, assigning higher spatial resolution to earlier ViT layers, and achieves the best overall accuracy.

\textbf{Ablation of Multi-Layer Feature Sources.} Table~\ref{tab:ablation_feature_source} studies the impact of feature inputs to the DPT-style head. Using only ViT features underperforms due to limited learnable parameters. LLM-only features improve results but remain suboptimal, as they lack fine-grained geometry details from early ViT layers. Combining multi-scale ViT features with the LLM final hidden state yields the best performance, effectively integrating high-level semantic information with detailed geometry cues.

\textbf{One-Stage vs. Two-Stage Training.}
We compare different training strategies in Table~\ref{tab:training_strategy}, including single-stage and two-stage variants. \textit{Stage-1 Only} preserves multimodal reasoning but yields limited depth gains. \textit{Stage-2 Only} improves geometry prediction, yet suffers from unstable optimization and reduced multimodal performance. The full two-stage strategy achieves a better balance. Unfreezing the vision encoder in Stage-2 further improves depth accuracy but harms general multimodal ability. Overall, our final design balances dense geometry prediction with the VLM's general capability.
\textbf{Effect of Focal-Length Normalization.}
We validate the focal-length normalization strategy in Table~\ref{tab:ablation_focal_length}. Training on raw mixed-source data suffers from camera-induced scale ambiguity, as identical scenes with different focal lengths yield inconsistent metric depth supervision. Normalizing inputs to a shared focal length $f_{\mathrm{c}}$ aligns projective geometry across datasets and mitigates this issue. We sweep $f_{\mathrm{c}}\!\in\!\{800,\,1000,\,1200\}$: smaller $f_{\mathrm{c}}$ loses image details, while larger $f_{\mathrm{c}}$ amplifies interpolation artifacts. Empirically, $f_{\mathrm{c}}\!=\!1000$ performs best across datasets, consistent with DepthLM~\cite{cai2025depthlm}.
 
\textbf{Inference Efficiency Analysis.}
We compare the efficiency of VLM-based methods in Table~\ref{tab:efficiency} using $256{\times}192$ inputs and reporting end-to-end runtime. DepthLM~\cite{cai2025depthlm} formulates depth estimation as per-pixel text queries, requiring $H{\times}W$ forward passes, making inference prohibitively slow. Youtu-VL~\cite{wei2026youtu-vl} reduces this to one pass but predicts sparse patches that require upsampling, introducing artifacts and overhead. In contrast, our method directly decodes multi-scale features into a pixel-aligned depth map in one pass without post-processing. This efficiency stems from treating the VLM as a native dense predictor, enabling efficient dense geometry prediction within a unified framework.

\section{Conclusion and Limitations}
In this paper, we present DepthVLM, a unified foundation model that jointly supports low-level dense geometry prediction and high-level multimodal understanding. We integrate a lightweight depth head into the VLM backbone and adopt a two-stage training strategy under the unified vision-text supervision, enabling geometry prediction and language response in a single forward pass. Extensive experiments show that DepthVLM achieves leading performance across diverse datasets with higher inference efficiency, surpasses strong pure vision models, and improves complex 3D spatial reasoning.\\
\noindent\textbf{Limitations.} This work mainly focuses on dense metric depth estimation and does not yet explore broader 3D perception tasks such as object detection and pose estimation. Extending the framework toward a unified model for holistic 3D perception and reasoning remains future work.





\clearpage
{\small
\bibliographystyle{plain}
\bibliography{reference}
}

\newpage
\appendix

\section*{Technical Appendices and Supplementary Material}

\section{Statistics of DepthVLM-Bench}
We summarize the data composition of DepthVLM-Bench in Tables~\ref{tab:train_dataset_stats} and~\ref{tab:benchmark_dataset_stats}. 


For training, we construct a large-scale and diverse dataset by sampling from the \emph{training splits} of multiple public benchmarks spanning both autonomous driving and indoor scenes, including Argoverse2~\cite{wilson2023argoverse}, Waymo~\cite{sun2020scalability}, DDAD~\cite{guizilini2020ddad}, NuScenes~\cite{caesar2020nuScenes}, ScanNet++~\cite{yeshwanth2023scannet++}, Taskonomy~\cite{zamir2018taskonomy}, HM3D~\cite{ramakrishnan2021hm3d}, and Matterport3D~\cite{chang2017matterport3d}. As many RGB images are extracted from videos and thus contain highly redundant, near-duplicate frames, we apply uniform sampling to reduce redundancy. Most datasets contribute approximately 800K images, resulting in a relatively balanced distribution across domains and helping mitigate dataset bias. Smaller datasets, such as DDAD and Matterport3D, are included at their original scales to further enrich diversity. In total, the training set comprises 4.4M images, covering a wide range of scenes, viewpoints, and depth distributions. We train for a single epoch. The 8B variant is trained for four days on 80 NVIDIA H20 GPUs, while the 4B variant is trained for two days using the same computational resources.

For evaluation, we assemble a comprehensive benchmark using the \emph{validation or test splits} of the same public datasets, covering both outdoor and indoor scenarios. When possible, each dataset contributes around 1K images to ensure a balanced and consistent evaluation protocol across sources. In addition, we include several standard indoor benchmarks, including ETH3D~\cite{schops2017eth3d}, sunRGBD~\cite{song2015sun}, IBims-1~\cite{koch2018ibims1}, and NYUv2~\cite{nyuv2}, to further assess generalization across diverse environments and capture varying depth characteristics. This design enables a fair and reliable evaluation of cross-domain generalization and robustness for dense geometry prediction.

\begin{table}[h]
\centering
\caption{\textbf{Training data statistics.} Number of training images sampled from each source dataset in our setting.}
\label{tab:train_dataset_stats}
\setlength{\tabcolsep}{4pt}
\renewcommand{\arraystretch}{1.1}
\resizebox{\linewidth}{!}{%
\begin{tabular}{lcccccccc}
\toprule
\textbf{Dataset} & Argoverse2 & Waymo & DDAD & NuScenes & ScanNet++ & Taskonomy & HM3D & Matterport3D \\
\midrule
\textbf{\# Images} & 800K & 800K & 76K & 206K & 800K & 800K & 800K & 158K \\
\bottomrule
\end{tabular}%
}
\end{table}

\begin{table}[h]
\centering
\caption{\textbf{Benchmark data statistics.} Number of evaluation images sampled from each dataset.}
\label{tab:benchmark_dataset_stats}
\setlength{\tabcolsep}{4pt}
\renewcommand{\arraystretch}{1.1}
\resizebox{\linewidth}{!}{%
\begin{tabular}{lccccccccc}
\toprule
\textbf{Dataset} & Argoverse2 & Waymo & DDAD & NuScenes & ETH3D & ScanNet++ & sunRGBD & IBims-1 & NYUv2 \\
\midrule
\textbf{\# Images} & 1K & 1K & 1K & 1K & 454 & 1K & 1K & 100 & 654 \\
\bottomrule
\end{tabular}%
}
\end{table}

\section{Evaluation of VLMs on Metric Depth Estimation}
\label{sec:appendix_vlm_eval}

To fairly compare the metric depth estimation capabilities of existing VLMs with our model, we design a standardized evaluation protocol described below.

\begin{figure}[t]
\centering
\includegraphics[width=1.0\linewidth]{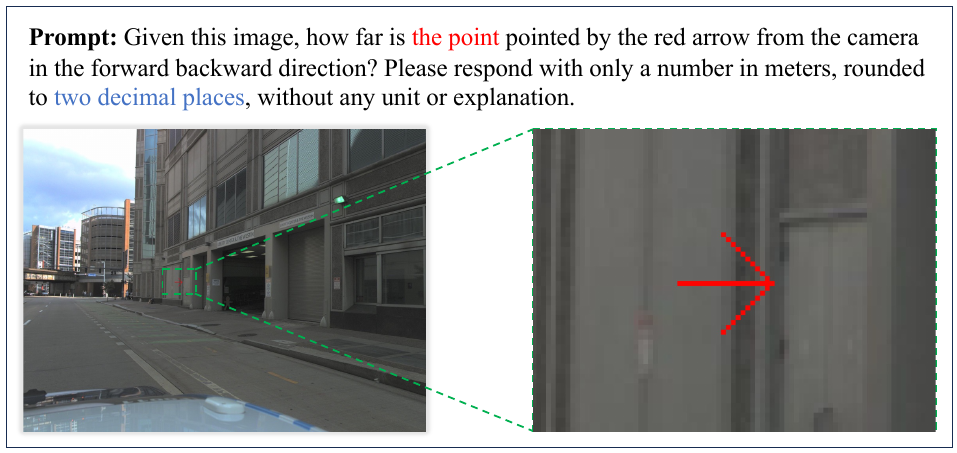}
\vspace{-1mm}
\caption{\textbf{VLM evaluation setup for metric depth estimation.} A red arrow marker (20 pixels) is drawn on the input image to indicate the query point. The model receives the annotated image along with a text prompt asking for the metric depth (in meters). The zoomed-in region shows the arrow marker in detail.}
\label{fig:test_vlm}
\vspace{-1mm}
\end{figure}

Following the paradigm introduced by DepthLM~\cite{cai2025depthlm}, given a single RGB image, a specific pixel location is indicated by a \textbf{red arrow marker} drawn directly on the image. The model is asked to estimate the metric depth (in meters), \emph{i.e.}, the distance along the camera's optical axis from that point to the camera, as illustrated in Figure~\ref{fig:test_vlm}.


\textbf{Visual Marker Size.}
DepthLM uses a default red arrow marker of \textbf{5 pixels}. However, through our experiments, we observe that off-the-shelf VLMs that have not been specifically trained for this task cannot reliably detect a 5 pixels marker. When not explicitly constrained to output only a number, these models frequently respond with statements such as:
\begin{quote}
\textit{``There is no red arrow in the image. Therefore, the distance cannot be determined.''}
\end{quote}
To ensure fairness and to maximally elicit each model's depth estimation capability, we increase the red arrow marker size to \textbf{20 pixels} for all evaluated VLMs. This larger marker is clearly visible in all tested scenes, ensuring that the evaluation measures depth estimation ability rather than marker detection ability.

\textbf{Image Resolution Handling.}
Since different datasets are captured at varying resolutions, we downscale images with a longest edge over 1024 pixels to this limit \emph{before} adding the red arrow marker, while leaving smaller images unchanged. This provides a consistent input condition across all models and datasets, and ensures that the marker remains clearly visible at the given image resolution.

With the above protocol, all VLMs are evaluated under identical conditions, ensuring a fair comparison with our method. For VLMs such as DepthLM~\cite{cai2025depthlm} and Youtu-VL~\cite{wei2026youtu-vl}, which are specifically trained for metric depth estimation, we directly adopt their provided evaluation scripts.

\section{Asset License and Consent}
\label{sec:asset license and consent}
Committed to openness and transparency to mitigate misinformation concerns~\citep{wei2024physical,wei2023moire}, we provide the licenses and URLs of all public datasets and benchmarks used in this work. These resources span a wide range of multimodal tasks~\cite{lu2022sqa,li2023pope,liu2023ocrbench,liu2024mmbench,fu2024blink}, including depth estimation, general multimodal understanding, spatial reasoning, document understanding, multi-image reasoning, visual grounding, and hallucination evaluation. Table~\ref{tab:License} summarizes all datasets and benchmarks used in this study along with their corresponding licenses and URLs.

\begin{table}[ht]
  \centering
  \caption{Open-source resources utilized in this paper.}
  \vspace{1mm}
  \renewcommand{\arraystretch}{1.2}
  \resizebox{\textwidth}{!}{%
    \begin{tabular}{lll}
    \toprule
    Name & License & URL \\
    \midrule
    ScienceQA & MIT License & \url{https://github.com/lupantech/ScienceQA} \\
    POPE &  MIT License  & \url{https://github.com/AoiDragon/POPE}  \\ 
    OCRBench & MIT License & \url{https://github.com/Yuliang-Liu/MultimodalOCR} \\
    MMBench & Apache License 2.0  &  \url{https://github.com/open-compass/MMBench} \\
    BLINK & Apache License 2.0 & \url{https://github.com/zeyofu/BLINK_Benchmark} \\
    Argoverse2    & Apache License 2.0 & \url{https://github.com/argoverse/argoverse-api} \\
    Waymo   & Apache License 2.0 & \url{https://github.com/waymo-research/waymo-open-dataset} \\
    NuScenes     & Apache License 2.0 & \url{https://github.com/nutonomy/nuscenes-devkit} \\
    
    MMStar & No license specified & \url{https://github.com/MMStar-Benchmark/MMStar} \\
    CountBenchQA & No license specified & \url{https://huggingface.co/datasets/vikhyatk/CountBenchQA} \\
    ScanNet++    &  ScanNet++ Terms of Use & \url{https://scannetpp.mlsg.cit.tum.de/scannetpp/} \\
    DDAD     & Creative Commons Attribution 4.0 & \url{https://github.com/TRI-ML/DDAD} \\
    ETH3D    & Creative Commons Attribution 4.0 & \url{https://www.eth3d.net/datasets} \\
    \bottomrule
    \end{tabular}%
    }
  \label{tab:License}%
\end{table}



\end{document}